%% file: main.tex
\documentclass[10pt,twocolumn,letterpaper]{article}

\usepackage[pagenumbers]{cvpr} 

\usepackage{graphicx}
\usepackage{amsmath}
\usepackage{amssymb}
\usepackage{booktabs}
\usepackage{amsthm,amsmath,amssymb}
\usepackage[ruled,vlined]{algorithm2e}
\usepackage{bm}
\usepackage{cite}
\usepackage{tabularx}
\usepackage{multirow}
\usepackage{algorithmic}
\usepackage{threeparttable}
\usepackage{float}
\usepackage{graphicx}

\usepackage[pagebackref,breaklinks,colorlinks,citecolor=cvprblue]{hyperref}
\usepackage[capitalize]{cleveref}
\crefname{section}{Sec.}{Secs.}
\Crefname{section}{Section}{Sections}
\Crefname{table}{Table}{Tables}
\crefname{table}{Tab.}{Tabs.}

\input{preamble}

\definecolor{cvprblue}{rgb}{0.21,0.49,0.74}

\def\paperID{1942} 
\def\confName{CVPR}
\def\confYear{2024}

\title{Episodic-free Task Selection for Few-shot Learning}

\author{Tao Zhang\\
Chengdu Techman Software Co., Ltd.\\
{zhangtao@alu.uestc.edu.cn}
}

\begin{document}
\maketitle  
\input{0_abstract} 
\input{1_intro}

\input{2_related}
\input{3_problem}
\input{4_methodology}
\input{5_experimental}

\input{6_discussion}
\input{7_conclusion}

{
    \small
    \bibliographystyle{ieeenat_fullname}
    \bibliography{main}
}


\end{document}

%% file: preamble.tex
\usepackage[dvipsnames]{xcolor}


%% file: 0_abstract.tex
\begin{abstract}
\label{sec:abstract}
Episodic training is a mainstream training strategy for few-shot learning. In few-shot scenarios, however, this strategy is often inferior to some non-episodic training strategy, e. g., Neighbourhood Component Analysis (NCA), which challenges the principle that training conditions must match testing conditions. Thus, a question is naturally asked: How to search for episodic-free tasks for better few-shot learning? In this work, we propose a novel meta-training framework beyond episodic training. In this framework, episodic tasks are not used directly for training, but for evaluating the effectiveness of some selected episodic-free tasks from a task set that are performed for training the meta learners. The selection criterion is designed with the affinity, which measures the degree to which loss decreases when executing the target tasks after training with the selected tasks. In experiments, the training task set contains some promising types, e. g., contrastive learning and classification, and the target few-shot tasks are achieved with the nearest centroid classifiers on the miniImageNet, tiered-ImageNet and CIFAR-FS datasets. The experimental results demonstrate the effectiveness of our approach.
\end{abstract}

%% file: 1_intro.tex
\section{Introduction}
\label{sec:intro}

The concept of few-shot learning (FSL) emerges as a solution to address the challenge of categorizing a limited number of samples that belong to previously unseen classes. \cite{fei2006one,wang2020generalizing}. FSL scenarios frequently arise in domains where sample acquisition proves challenging, such as drug discovery \cite{altae2017low}, agriculture \cite{yang2022survey} and healthcare \cite{leung2021health}, thereby receiving increasing attention. A widely used framework for FSL is meta-learning, which follows the principle that ``test and train conditions should match'' \cite{vinyals2016matching} or ``the process of improving a learning algorithm
over multiple learning episodes''\cite{hospedales2021meta}, aiming to learn transferable knowledge for inference.

\begin{figure}[h]
	\centering
	{\includegraphics[width=0.45\textwidth]{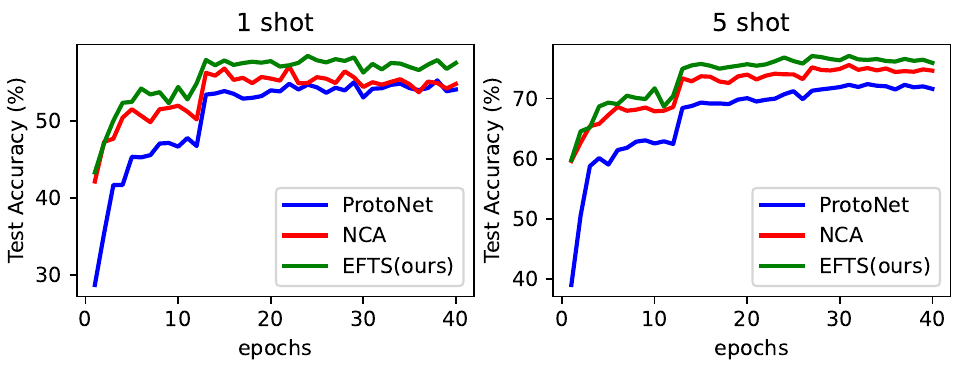}}
	\caption[]{Test accuracies obtained by using Prototypical Network (ProtoNet), Neighbourhood Component Analysis (NCA) and our proposed Episodic-free Task Selection (EFTS) for 5-way 1-shot (left) and 5-way 5-shot (right) tasks on \textit{mini}ImageNet, respectively. In target tasks, ProtoNet is used as the classifiers.}
	\label{problem}
\end{figure}

Meta-learning is endeavoring to address the FSL problem through episodic training, where a batch of samples is divided into a support set and a query set to mimic the test conditions \cite{vinyals2016matching}. Recently, numerous FSL methods have been proposed based on the episodic training \cite{vinyals2016matching, snell2017prototypical, simon2020adaptive, lee2019meta, finn2017model,allen2019infinite, oreshkin2018tadam,sung2018learning, yang2022few}, in which base-learners are trained in an inner loop, and meta-learners are trained in an outer loop. Furthermore, some work shows that the meta-learning models can still maintain good performance even after omitting the inner loop, e. g., NIL \cite{tian2020rethinking}.

Here raises a question: \textit{is the most matching task always the best training task?} Some recent experimental results shows that episodic training is often inferior to non-episodic training in few-shot learning, especially in the cross-domain scenarios \cite{tian2020rethinking, chen2019closer, laenen2021episodes, chen2020closer, raghu2019rapid}. To show this, we made an illustration in Fig. \ref{problem}.  Figure \ref{problem} shows that the non-episodic training (NCA) outperforms the episodic training (ProtoNet). The reason can be analyzed from different perspectives. Firstly, episodic training discards a large amount of available sample information that is helpful for model training \cite{laenen2021episodes}. Secondly, the learned embedding representation in meta-learning is still not good enough \cite{tian2020rethinking, raghu2019rapid, du2020few}.  

\begin{figure*}[h]
	\centering
	{\includegraphics[width=0.95\textwidth]{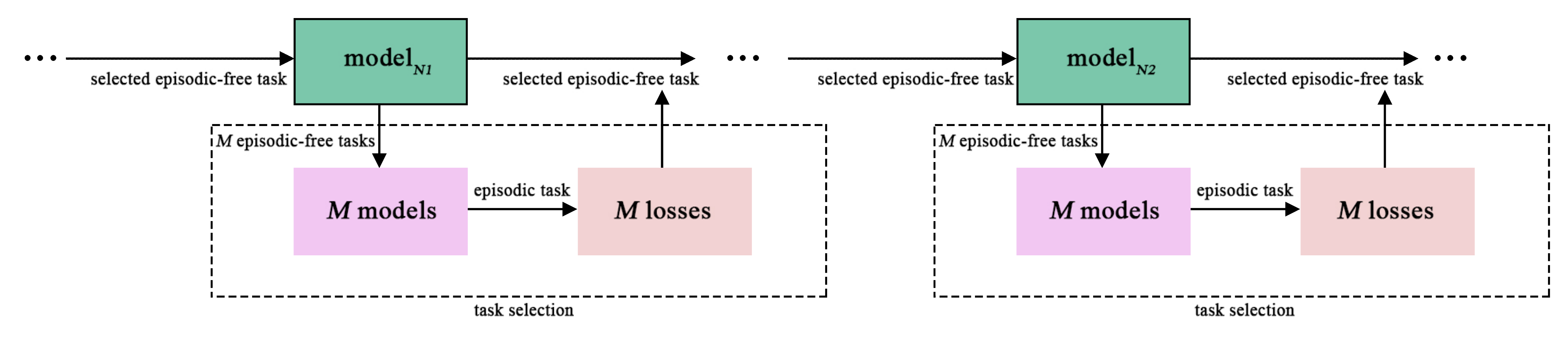}}
	\caption[] {The framework of the proposed EFTS. In this task selection section, firstly, using M episodic-free tasks updates the model respectively; Secondly, M updated models are evaluated using the episodic task, and thus M losses are obtained; finally, M losses are compared, and the episodic-free task corresponds to the smallest loss will be selected for the next stage of training. Task selection can be done multiple times during the training process.}
	\label{Ablation0}
\end{figure*} 

In this work, we address the problem of how to design a framework for selecting the better training tasks for few shot learning. Unlike episodic training, the episodic tasks (episodes in \cite{laenen2021episodes}) that are matching to the target tasks are not applied directly for training in our framework, but as a proxy for evaluating the benefits of the episodic-free tasks (episodes or not episodes) to the target tasks. After this evaluation, the optimal tasks are selected as the task for training the meta-learners (Fig. \ref{Ablation0}).   

To the best of our knowledge, the mutual influence between tasks that have been studied widely in multi-task learning \cite{fifty2021efficiently,upadhyay2023multi,sener2018multi} are rarely applied for few-shot learning. In this paper, a simple and effective approach for inter-task evaluation in multi-task learning, the affinity \cite{fifty2021efficiently}, will be used in our work as a task selection criterion, which is called as episodic-free task selection (EFTS). In EFTS, the tasks that have the highest affinity scores are selected to jointly train the meta-learners, where any ingenious multi-task learning strategies are not adopted. Figure \ref{problem} shows that, compared to the other two single-task fixed  strategies, EFTS as a multi-task selection strategy can perform the best.

The contributions of this paper are summarized:

1. A novel framework for few-shot learning is proposed. In this framework, the training tasks can be episodic-free, and their effectiveness can be evaluated by a series of episodic tasks.

2. A criterion called affinity that is popular in multi-task learning is introduced into few-shot learning for task selection, by which the effective single tasks that have the highest affinity scores can be selected for jointly training the meta-learners.

3. A series of experiments focusing on the effectiveness of the proposed framework have been conducted, and the influence of hyper-parameters have been discussed.

%% file: 2_related.tex
\section{Related Work}
\label{sec:Related}
In this section, we review the research in multi-task learning, few-shot learning and the cross-interactions between them. The matching principle in meta-learning provides a relationship between source and target tasks, which requires that the source  task should be episodic. In an episodic task,  a batch of samples is divided into support and query, aiming to mimic the test conditions \cite{vinyals2016matching}. The episodic training as a standard method is applied in a lot of few-shot approaches \cite{vinyals2016matching, snell2017prototypical, finn2017model, oreshkin2018tadam, simon2020adaptive}. Oriol Vinyals et al. proposed Matching Nets for few-shot learning , whose training procedure obeys that test conditions must match train conditions \cite{vinyals2016matching}. Based on Matching Nets, Jake Snell et al. proposed prototypical networks (ProtoNet) for few-shot learning \cite{snell2017prototypical}. In this approach, a class is represented as a prototype, which is calculated as an average of embeddings of support samples belonging to this class. Subsequently, various improvements are constantly emerging for ProtoNet \cite{pahde2021multimodal, fort2017gaussian, allen2019infinite, oreshkin2018tadam, liu2020prototype}. Flood Sung. et al. proposed the Relation Network to learn a deep distance metric for comparing the samples in the episodes \cite{sung2018learning}. Chelsea Finn et al. proposed the Model-Agnostic Meta-Learning (MAML), which trains the base learner in an inner loop for achieve rapid adaptation, and trains the meta learner in an outer loop \cite{finn2017model}.

These elaborate episodic training based few-shot approaches are questioned recently because in a cross-domain scenario, their performances are inferior to the simple baselines that are often used for transfer learning \cite{chen2019closer}. In this work by Steinar Laenen et al. \cite{laenen2021episodes}, a series of systematic experiments were conducted to compare ProtoNet and Neighbourhood Component Analysis (NCA), and it was found that NCA outperformed ProtoNet because NCA was able to fully utilize the information about the similarity between samples, while this information is largely discarded in ProtoNet. In addition, Yinbo Chen et al. designed  the Meta-Baseline that combines meta-learning and whole classification for few-shot learning, by which they found meta-learning model that performs better in generalization on unseen tasks from base classes may exhibit poorer performance on tasks from novel classes \cite{chen2021meta}. The essence of this problem may be related to representations. Yonglong Tian et al. proposed a baseline showing that learning a good representation is important in few-shot learning, even with just a simple linear classifier \cite{tian2020rethinking}. These work provides the straightforward presentation of the key problems in episodic training and a series of effective analysis, which may inspire a novel few-shot framework to improve episodic training.

Multi-task learning leverages the transfer of information between tasks to provide inductive biases for the target task. In multi-task learning, the weighting of task loss will greatly affect the performance of the model. In \cite{kendall2018multi}, the weighting problem is solved by a principled approach using homoscedastic uncertainty. In addition, there are inter-dependencies between tasks, and thus multi-task learning problems can be considered as multi-objective optimization problems and require a Pareto optimal solution \cite{sener2018multi}. In \cite{fifty2021efficiently}, an intuitive evaluation method called inter-task affinity was proposed, which evaluates how well the model performs in a task when completing another task. This method is used for task grouping training. Recently, the multi-task meta learning (MTML) that combines MTL and bi-level meta optimization was proposed, which introduces the multi-task learning episodes  architecture into MTL scenarios \cite{upadhyay2023multi}.  MTML can result in an ensemble that can achieve better performance for an unseen task in fewer steps compared to training the task individually from scratch. 

Multi-task learning approaches may provide a good representation for few-shot learning \cite{bouniot2022improving, du2020few}. \cite{du2020few} proposed a theoretical analysis of learning a common good representation between source and target task with multi-learning, aiming to reveal the maximum extent of sample size reduction. \cite{bouniot2022improving} explores the framework of multi-task representation (MTR) learning, which aims to leverage source tasks to acquire a representation that reduces the sample complexity of solving a target task.  The aforementioned work points out that a representation is crucial for connecting the source and target tasks, but not provides an approach to directly evaluate the benefits of a source task to a target few-shot task.

%% file: 3_problem.tex
\section{Problem Definition}
Let us consider a set of $n$ meta-training tasks $\mathcal{T}_{t}=\{\tau_1, \tau_2,...,\tau_T\}$ that can be any types, not limited to just episodes or whole-classification. The $i$th meta-training task $\tau_i$ denotes the $i$th task set that contains a sample set $\mathcal{X}$. In addition, consider that a meta-training task is optimized using a set of the $N$-way $K$-shot meta-evaluation tasks $\mathcal{T}_{e}=\{\nu_1, \nu_2,...,\nu_L\}$ that maintain consistency with the conditions of the target few-shot tasks. The $l$th $N$-way $K$-shot meta-evaluation task $\nu_l$ with arbitrary $l$ contains the support set $\mathcal{S}_l$ and the query set $\mathcal{Q}_l$. The aim is achieving the target few-shot tasks using the meta-learners that are trained by $\mathcal{T}_{t}$.

%% file: 4_methodology.tex
\section{Methodology}
\subsection{Preliminary}
\textbf{Episodic Training.} A few-shot learning (FSL) model is commonly trained by completing a series of episodic tasks on a few-data scenario that is similar to those in test stage \cite{vinyals2016matching}. In each task, a few data is randomly collected for training meta-learners and base-learners in this model, which contains  support set $\mathcal{S}= \{(\bm{s}_1,y_1),(\bm{s}_2,y_2),...,(\bm{s}_N,y_N)\}$ and  query set $\mathcal{Q}=\{(\bm{q}_1,y_1),(\bm{q}_2,y_2),...,(\bm{q}_M,y_M)\}$. In the training stage, the base-learners are commonly optimized  with $\mathcal{S}$ in the inner loop, aiming to rapidly adapt to a new task, and the meta-learners are iteratively optimized in the outer loop with $\mathcal{Q}$ in classification, or $\mathcal{S}$  and $\mathcal{Q}$ in metric learning.  The meta-learner training thus is the main considerations in few-shot learning, and the base-learner training in the inner loop in some cases, e. g., the metric-based meta-learning, can be omitted.

\textbf{Whole-classification Training.} Meta-learner training by completing the episodic tasks is only considering the connections between $\mathcal{S}$ and $\mathcal{Q}$, and massive internal connections in them are discarded. These inadequate utilization of information would degrades the performance of the FSL model. Thus, the whole-classification training strategy is proposed for addressing this issue \cite{laenen2021episodes}. In this strategy, a batch is not divided into support and query set, but as a whole for training the meta-learners. In metric-based meta learning, this strategy can consider all the connections in the batch, which may benefit for the model training.

\textbf{Inter-Task Affinity.} Hard parameter sharing is a way for multi-task learning that involves sharing the hidden layers among all tasks while maintaining separate task-specific output layers. During the training process, information is transmitted between tasks through the gradients of parameters in these shared hidden layers. The degree of impact between tasks can measured by inter task affinity \cite{fifty2021efficiently}. Inter task affinity provides an effective way to cluster tasks for model training.

\subsection{Episodic-free Task Selection (EFTS)}
In general, the episodic-free tasks can be homogeneous or heterogeneous. Motivated by the work \cite{fifty2021efficiently} that groups the tasks via inter-task affinity that measures the impact of a task’s gradient updates on the objective of other tasks, we can select meta-training tasks by evaluating their gradient updates impacting the $N$-way $K$-shot meta-evaluation tasks. Suppose that a meta-training task is embodied with a subset $\mathcal{T}_{sub}$ of $\mathcal{T}_{set}=\{t_1,t_2,...,t_S\}$ where $\mathcal{T}_{set}$ is the set of given tasks and $t_i$ denotes a single task. Consider a batch of samples $\mathcal{X}$ inputs a network $f_\theta$ with parameters $\theta = \{\theta_{s},\theta_{i}\}$ with $\theta_{s}$ the shared parameters that is shared by all the tasks, and $\theta_{i}$ the specific parameters that is proprietary in different tasks. The loss function with input $\mathcal{X}_t$ for the task set $\mathcal{T}_{sub}$ in a $t$th update by gradient descent can be expressed as 
\begin{equation}\label{update_m1}
	\theta^{t+1}_{s} = \theta^{t}_{s} - \lambda\nabla_{\theta_s}\mathcal{L}_{\mathcal{T}_{sub}}(\mathcal{X}^t;\theta_{s},\theta_{i}^t),
\end{equation}
where $\lambda$ is the learning rate that is a hyper-parameter for training. It should be noted that $\mathcal{L}_{\mathcal{T}_{sub}}$ in Eqs. \ref{update_m1} can be implemented with the approaches of multi-task learning. Enlightened by the work \cite{fifty2021efficiently}, we can introduce the affinity from $\mathcal{T}_{sub}$ to the target tasks $\mathcal{T}_{e}$: 
\begin{equation}\label{affinity1}
	\mathcal{Z}_{\mathcal{T}_{sub} \to \mathcal{T}_{e}}^{t,\rm{UNA}} = 1 - \frac{\sum_{l=1}^L\mathcal{L}_{\nu_l}(\mathcal{S}^l,\mathcal{Q}^l;\theta^{t+\rm{UNA}}_{s},\theta_{i})}{\sum_{l=1}^L\mathcal{L}_{\nu_l}(\mathcal{S}^l,\mathcal{Q}^l;\theta^{t}_{s},\theta_{i})},	
\end{equation}
where UNA is the update number per affinity. In Eq. \ref{affinity1}, as $\mathcal{Z}_{\mathcal{T}_{sub} \to \mathcal{T}_{e}}$ is larger, the selected task set in $t$th has more positive impacts on the target tasks $\mathcal{T}_{e}$ in the $t$th update. Here we expect the model to acquire a transferable knowledge, and thus the $\mathcal{S}^l$ and $\mathcal{Q}^l$ in Eq. \ref{affinity1} are different with  $\mathcal{X}_t$ as possible. This implies that $\mathcal{T}_{sub}$ may not have a high affinity with $\mathcal{T}_{e}$, even they are implemented with the same classifiers (e. g., ProtoNet). In addition, \cite{fifty2021efficiently} shows that the affinity is calculated across  multiple steps  is better than single step. Thus, here we propose the averaged affinity across all UNA multiple steps, i.e,
\begin{equation}\label{affinity2}
	\hat{\mathcal{Z}}^{t, \rm{UNA}, M} _{\mathcal{T}_{sub}\to \mathcal{T}_{e}} = \frac{1}{M}\sum_{i=1}^{M} \mathcal{Z}^{t + (i-1) \times \rm{UNA}, \rm{UNA}}_{\mathcal{T}_{sub} \to \mathcal{T}_{e}}.	
\end{equation}
Then, we can use the average affinity in Eq. \ref{affinity2} as a criterion to select the optimal $\mathcal{T}_{sub}^*$ in an update, i. e., 
\begin{equation}\label{sele_min}
	\mathcal{T}_{sub}^* = arg\min_{\mathcal{T}_{sub} \subset \mathcal{T}_{set} }\hat{\mathcal{Z}}^{t, \rm{UNA}, M} _{\mathcal{T}_{sub}\to \mathcal{T}_{e}}.	
\end{equation}

Equation \ref{sele_min}  requires solving a 0-1 integer programming problem that would be NP-hard. When the total of optional meta-training tasks is small, we can use the exponential search strategy such as exhaustive search and branch and bound to select the optimal task subset. But as it is large, it is computationally expensive. In this case, it can be solved by some other strategies, e. g.,  sequential algorithms and randomized algorithms, which are often used as the wrapper approaches for feature selection \cite{agrawal2021metaheuristic, awadallah2020binary}. Here we simply calculate the loss of a sub-optimal $\mathcal{T}_{sub}^*$, or called the EFTS objective function:
\begin{equation}\label{sele_min_single}
	\mathcal{L}_{\mathcal{T}_{sub}^*}(\mathcal{X}^{t};\theta_{s},\theta_{i}^{t}) = \sum_{i=1}^Q 	\mathcal{L}_{\tau_i*}(\mathcal{X}^{t};\theta_{s},\theta_{i}^{t}),	
\end{equation}
where $\mathcal{L}_{\tau_i*}$ belongs to the set that contains the $Q$ single tasks from $\mathcal{T}_{t}$ that have the highest $Q$ affinity scores calculated in Eq. \ref{affinity2}.  After the stage of task selection, the few-shot models are training with

\begin{equation}\label{update_final_m1}
	\theta^{t+1}_{s} = \theta^{t}_{s} - \lambda\nabla_{\theta_s}\mathcal{L}_{\mathcal{T}_{sub}^*}(\mathcal{X}^{t};\theta_{s},\theta_{i}^{t}).
\end{equation}
It should be noted that EFTS can be applied in any update during the entire training process and can be applied multiple times. The pseudo-code for the proposed EFTS is shown in Algorithm \ref{algMETAS}.

\begin{algorithm}[!h]
	\caption{EFTS for Few-shot Learning}
	\label{algMETAS}
	\renewcommand{\algorithmicrequire}{\textbf{Input:}}
	\renewcommand{\algorithmicensure}{\textbf{Output:}}
	\begin{algorithmic}[1]
		\REQUIRE Training set $\mathcal{D}$, evaluation set $\mathcal{V}$, learning rate $\lambda$, maximum number of iterations $Maxitr$,  and number of evaluation tasks $L$, time-point set of task selection $\Theta$  
		\ENSURE $\theta^{Maxitr}_i$, $\theta^{Maxitr}_s$   
		\STATE  randomly initialize the parameters $\theta_s$ and $\theta_i$
		\STATE  randomly initialize the meta-training task ${\mathcal{T}_{sub}^*}$ 
		\FOR{$t$ in $\{1,2,...,Maxitr\}$}
		\STATE Randomly sampling $\mathcal{X}^t$ from $\mathcal{D}$
		\ENDFOR	
		\FOR{$l$ in $\{1,2,...,L\}$}
		\STATE Randomly sampling $\mathcal{S}^l$ and $\mathcal{Q}^l$ from $\mathcal{V}$
		\ENDFOR
		\FOR{$t$ in $\{1,2,...,Maxitr\}$}
		\IF {$t \in \Theta$}
		\STATE Calculate the average affinity $\hat{\mathcal{Z}}^{t, \rm{UNA}, M} _{\mathcal{T}_{sub}\to \mathcal{T}_{e}}$ for each single task using Eqs. \ref{affinity1} and  \ref{affinity2}
		\STATE Calculate $\mathcal{L}_{\mathcal{T}_{sub}^*}(\mathcal{X}^{t};\theta_{s},\theta_{i}^{t})$ using Eq. \ref{sele_min_single} 
		\ENDIF
		\STATE $\theta^{t+1}_{s} = \theta^{t}_{s} - \lambda\nabla_{\theta_s}\mathcal{L}_{\mathcal{T}_{sub}^*}(\mathcal{X}^t;\theta_{s},\theta_{i}^t)$
		\ENDFOR 		
		\RETURN $\theta^{Maxitr}_i$, $\theta^{Maxitr}_s$
	\end{algorithmic}
\end{algorithm}

%% file: 5_experimental.tex
\section{Experimental Setup}
\subsection{Dataset}

\textbf{ \textit{mini}ImageNet.} The  \textit{mini}ImageNet dataset \cite{vinyals2016matching} consists of 84×84 pixel images sourced from ILSVRC-2012. It comprises 100 classes, with each class containing 600 images. The class division for meta-training, meta-validation, and meta-testing involves randomly splitting them into 64, 16, and 20 classes, respectively. This dataset is commonly used in the field of few-shot learning research.

\textbf{ \textit{tiered}-ImageNet.} The  \textit{tiered}-ImageNet dataset \cite{ren2018meta} also originates from ILSVRC-2012 and contains 84×84 pixel images. In comparison to  \textit{mini}ImageNet, it offers a more extensive variety, encompassing 608 classes. These classes are further divided into 34 high-level categories, with each category housing 10 to 30 classes. The  \textit{tiered}-ImageNet dataset is widely utilized for evaluating and benchmarking few-shot learning methods. All classes are divided into 351, 97 and 160 classes for meta-training, meta-validation, and meta-testing, respectively.

\textbf{CIFAR-FS.} The CIFAR-FS dataset \cite{bertinetto2018meta} comprises 32×32 pixel images extracted from CIFAR-100 \cite{krizhevsky2009learning}. It consists of 100 classes, with each class containing 600 images. CIFAR-FS is frequently employed in few-shot learning experiments as a valuable resource for evaluating the performance of various approaches. All classes are randomly divided into 64, 16 and 20 for meta-training, meta-validation, and meta-testing, respectively.

\subsection{Task Setting for EFTS}

In EFTS, a task set need to be designed for task selection. To ensure that our task settings can well simulate the situation of unknown tasks in practice, we do not expect each single task in this task set is promising. Here ten training tasks are used for EFTS: ProtoNet \cite{snell2017prototypical} (task1),  Neighbourhood Component Analysis (NCA) \cite{laenen2021episodes} (task2), Classification (task3, task4), and supervised contrastive learning (SupCon) with different augmentation strategy \cite{khosla2020supervised} (task5-task10). For the classification tasks, the number of classes is the same as the number of features in task4, and in task5, we set the number of classes in line with \cite{tian2020rethinking}. For task5-task10, there uses 6 pairs of  combinations between original samples and augmented samples with cropping, color distortion, Cutout, horizontal flipping. 

We mainly consider that the data sampling strategy that all tasks share the same batch of data during each update process, and this data is collected in episodes that can be divided into a support set and a query set. For episodic tasks, these episodes can be applied directly; For non-episodic tasks, we merge the support set and query set into a single batch for usage. Like \cite{laenen2021episodes}, the batchsize = (M + N) $\times$ C, where C is the way in every batch. For example, when the training way is 16, and the number support samples and query samples in each class is 5 and 3, respectively, the batchsize is 128. This data sampling strategy is not completely episodic-free in practice because data collection imitates those of target tasks, even though the classifiers are arbitrary. However, in order to have all tasks adopt the same inputs, we still consider it as our primary experimental approach. We still consider the strategy that all tasks share different data inputs: batchsize without considering $\{$M, N, C$\}$ for non-episodic tasks, and episodes for episodic tasks.

\subsection{Head Setting for Evaluation}

 In target tasks, the selection of the head significantly affects the performance of the model \cite{laenen2021episodes, tian2020rethinking}. Here are multiple ways to configure evaluation task for few-shot classification, e. g., KNN, nearest centroid and soft assignments \cite{laenen2021episodes}. In line with \cite{laenen2021episodes}, unless otherwise specified, ProtoNet as the default classifier will be used for the evaluation in this work. In addition, we considered using logistic regression (LR) \cite{tian2020rethinking} as an additional head for comparison.

\subsection{Implementation}
In training stage, the inputs contains training and validation sets from three datasets. For ProtoNet, the embeddings of training and evaluation samples are centred and normalised for all the tasks. Specifically, the centralization is implemented by $\bm{s}_i \leftarrow \bm{s}_i - \bm{\overline{s}}$ where $\overline{\bm{s}} = 1 / |\mathcal{S}| \sum_j \bm{s}_j $ for support samples, and  $\bm{q}_i \leftarrow \bm{q}_i - \overline{\bm{s}}$ for query samples. The normalization for all samples $\bm{x}$ is implemented by $\bm{x} \leftarrow \bm{x}/ ||\bm{x}||$. For LR head, the embeddings of training and evaluation samples are normalised for task1,task2 and task5-task10.

All the single tasks utilize ResNet-12 as the backbone, where the SGD optimizer is applied for the training. For the input sharing strategy, the learning rate is initially set as 0.1 decayed with a factor of 0.0025, 0.00032, respectively; For the strategy of not sharing inputs, the learning rate is initially set as 0.05 then decayed with a factor of 0.1. The processes of training and test are implemented in the PyTorch machine learning package \cite{paszke2017automatic}.

\begin{figure*}[h]
	\centering
	{\includegraphics[width=1\textwidth]{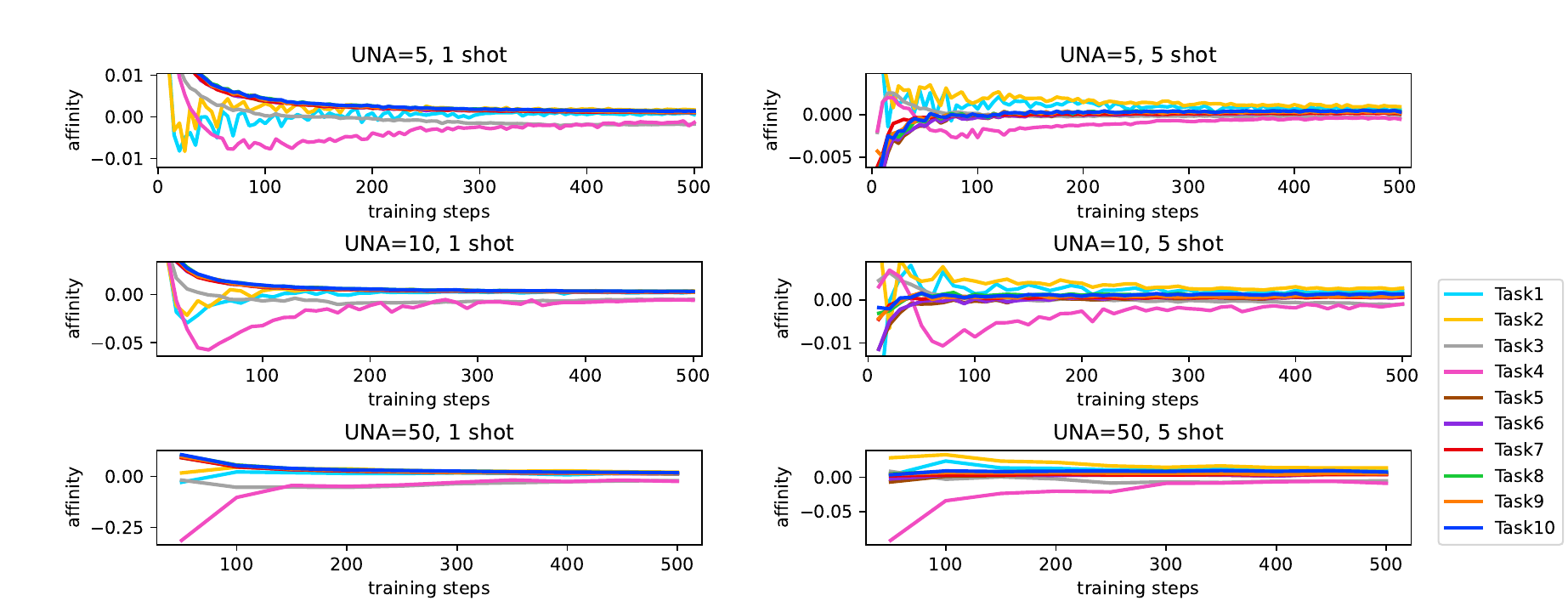}}
	\caption[]{Average affinity against training steps with different Update Number per Affinity (UNA) on CIFAR-FS using ResNet-12. The batchsize is 128 and the training way is 16.}
	\label{affinity}
\end{figure*}

\section{Experimental Results}

\subsection{Hyperparameter analysis}

{\bf{Impacts of update number per affinity.}} The impacts of update number per affinity (UNA) on the affinities of 10 tasks within 500 training steps after selecting the tasks at the beginning of training process are shown in Fig. \ref{affinity}.  128 batchsize and 16 way are applied in the model whose architecture is ResNet-12. As UNA is 5 or 10, the sequence of affinities pertaining to the task remains unsteady when the number M of affinity for averaging is small. As M increases, the affinity curves for UNA values at 5, 10, and 50 begin to stabilize progressively. In addition, as UNA increases, affinities become more stable while the step increases. For all of the UNA, the affinity relationship between tasks is relatively consistent. For example, overall both of task 3 and task 4 always have the lower affinities. In particular, we can see that the affinity of task1 (ProtoNet) is not the highest one, even in the evaluation tasks ProtoNet is also used as the classifier. 

\begin{figure}[htbp]
	\centering
	{\includegraphics[width=0.5\textwidth]{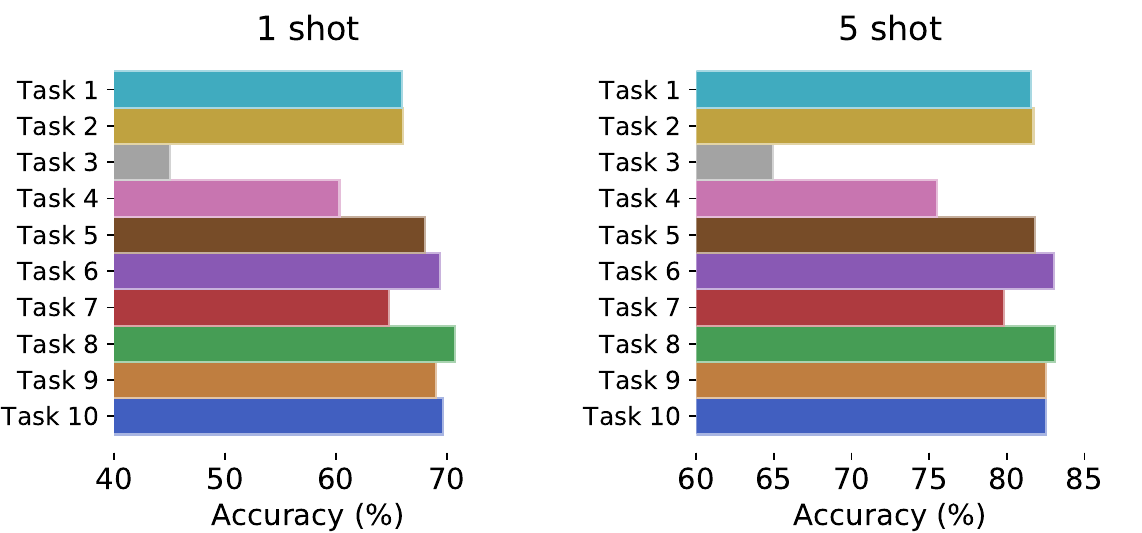}}
	\caption[]{Average affinity against training steps with different Update Number per Affinity (UNA) on CIFAR-FS using ResNet-12. The batchsize is 128 and the training way is 16.}
	\label{affinity11}
\end{figure}

Furthermore, to illustrate the effectiveness our methods, we report the performances of single tasks for reference, as shown in Fig. \ref{affinity11}. We can see that task 3, task 4 and task 7 perform the worst, and they also have the smallest affinities (see Fig. \ref{affinity}).  In addition, task5-task10 perform the best, and for 1 shot tasks, they can correspond well to their affinities (see Fig. \ref{affinity}). However, task3 and task4 are inferior to task5, task6 and task8-task10 for 5 shot tasks,  but have the highest affinities (right part of Fig. \ref{affinity}), probably because they converge quickly at the beginning of training.  

{\bf{Impacts of data sampling strategy.}} In multi-task learning scenarios that involve both episodic tasks and non-episodic tasks, we may adopt different data sampling strategies: 1) sharing a common input, i.e., episodes, for both episodic and non-episodic tasks; 2) using episodes as input for episodic tasks and using batches, without considering way, shot, and query, as input for non-episodic tasks. Additionally, the parameters of way, shot, and query can be set as hyper-parameters in an episode. For comparison convenience, we have fixed the batch size and compared these two strategies, and in episodes, we have fixed the sum of shot and query, as shown in Table \ref{numbers}. Table \ref{numbers} shows that, the sum of shot and query can effect the performance of EFTS while the inputs of all tasks are episodes (inputs A). For example, when the batchsize is 256, the maximum difference in strategy inputs A can exceed 2$\%$. As the inputs B is applied, the sum of shot and query has no significant impact on performance. Comparing two strategies, the inputs A may perform the best but depend on the setting of hyper-parameters, while the inputs B is still competitive and is almost unaffected by hyper-parameters.

\begin{figure*}[h]
	\centering
	{\includegraphics[width=1\textwidth]{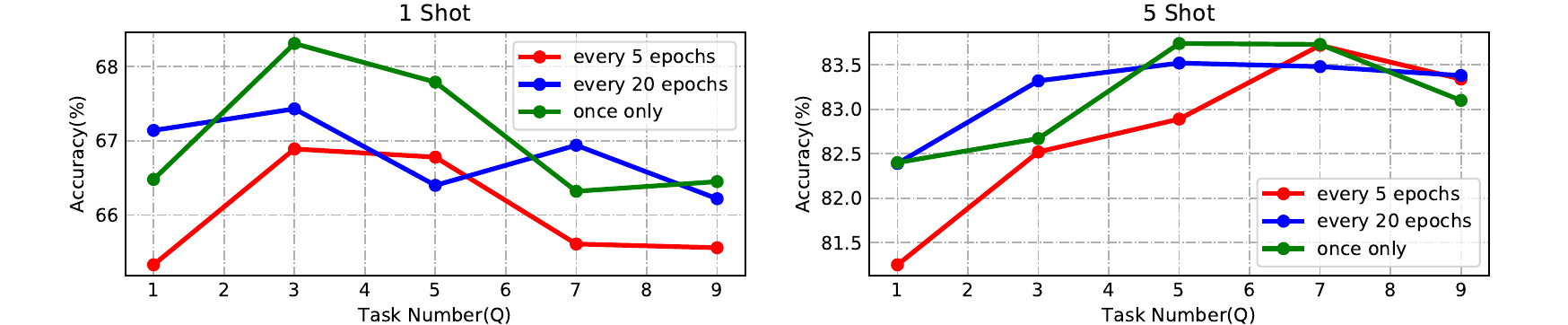}}
	\caption[]{Accuracies ($\%$) obtained by EFTS with different interval of task selection against different Q on CIFAR-FS using ResNet-12. The batchsize is 128 and the training way is 16. In addition, NUA = 50 and M = 4.}
	\label{AccEpoch2}
\end{figure*}

{\bf{Impacts of interval of task selection against Q.}} The interval between task selection with different number of jointly used tasks is discussed, as listed in Fig. \ref{AccEpoch2}.  We can see that for both of 1 shot and 5 shot tasks, frequent selection of tasks during the training process does not improve the performance of the model, Selecting tasks only once at the beginning of training is enough. In addition, when the task is selected only once, the accuracy increases from Q = 1 to Q = 5, and decreases from Q = 5 to Q = 9. The reason may be that, with small Q, some effective tasks are abandoned (e. g., task 8 and task 10); With large Q, some laggard tasks are included (e. g., task 3 and task 4). These results imply that an appropriate Q ensures a small fault tolerance rate and leverage the advantages of multitasking learning.

\subsection{Ablation Study}
In order to demonstrate that the proposed EFTS can effectively select the training tasks, we conduct ablation experiments in this section. Firstly, we compare the performances of ProtoNet, EFTS and all task selection with different batchsize for 1 shot and 5 shot tasks, as listed in Table \ref{batchsize}. The results show that EFTS at all cases outperforms the other two methods. The strategy for selecting all tasks may be simple, but it may introduce some tasks that have negative impacts on the model performance. In addition, the results supports the viewpoint that the best training task can be episodic-free, which is consistent with the results in \cite{laenen2021episodes}.

\begin{table}[h]
	\centering
	\begin{threeparttable}
		\caption{Test accuracies ($\%$) obtained by EFTS with different batchsize, training shot (ts) and training query (tq) for episodic tasks on CIFAR-FS, where Q = 5, UNA = 50, M = 4 and tasks are selected once only. A: All tasks sharing the same data inputs; B: Non-episodic tasks share different data inputs with episodic tasks, without considering ts and tq.  In addition, NUA = 50 and M = 4.}
		\begin{tabular}{cccccc}
			\toprule[1pt]
			\textbf{}
			{\bf{batchsize}}&\bf{inputs}&\bf{ts+tq}&\bf{1 shot}&\bf{5 shot}\\
			\hline
\multirow{6}{*}{128}   &A&8 &$67.79_{\pm 0.96}$ &$83.74_{\pm 0.65}$\\
		                    &A&16&$68.92_{\pm 0.94}$ &$83.86_{\pm 0.68}$\\
	                    	&A&32&$65.80_{\pm 0.97}$ &$82.82_{\pm 0.71}$\\
	                    	&B&8 &$68.23_{\pm 0.95}$ &$83.09_{\pm 0.64}$\\
	                    	&B&16 &$68.12_{\pm 0.94}$ &$83.23_{\pm 0.67}$\\
	                    	&B&32 &$68.08_{\pm 0.95}$ &$83.04_{\pm 0.64}$\\
	                    	\hline
\multirow{6}{*}{256}   &A&8 &$67.90_{\pm 0.98}$ &$82.48_{\pm 0.71}$\\
		                    &A&16&$69.18_{\pm 0.94}$ &$84.34_{\pm 0.68}$\\
                            &A&32&$69.40_{\pm 0.94}$ &$84.70_{\pm 0.67}$\\
                            &B&8 &$69.34_{\pm 0.95}$ &$84.21_{\pm 0.66}$\\
                            &B&16 &$69.96_{\pm 0.97}$ &$84.14_{\pm 0.63}$\\
                            &B&32 &$69.55_{\pm 0.94}$ &$84.23_{\pm 0.65}$\\
			\bottomrule[1pt]
		\end{tabular}\label{numbers}
	\end{threeparttable}
\end{table}

\begin{table}[h]
	\centering
	\begin{threeparttable}
		\caption{Comparison of accuracies ($\%$) obtained by three methods with different batchsize on CIFAR-FS.  In addition, NUA = 50 and M = 4.}
		\begin{tabular}{cccc}
			\toprule[1pt]
			\textbf{}
			\bf{batchsize}        &{\bf{method}}&\bf{1 shot}&\bf{5 shot}\\
			\hline
			\multirow{3}{*}{128}   &ProtoNet	 &$65.92_{\pm 0.97}$ &$81.56_{\pm 0.72}$\\
			&EFTS	     &$68.31_{\pm 0.98}$ &$83.74_{\pm 0.65}$\\
			&all tasks	 &$67.04_{\pm 0.99}$ &$83.10_{\pm 0.70}$\\
			\hline
			\multirow{3}{*}{256}  &ProtoNet	 &$67.39_{\pm 0.93}$ &$83.20_{\pm 0.69}$\\
			&EFTS	      &$69.40_{\pm 0.94}$ &$84.70_{\pm 0.67}$\\
			&all tasks	 &$66.25_{\pm 0.94}$ &$81.75_{\pm 0.74}$\\
			\bottomrule[1pt]
		\end{tabular}\label{batchsize}
	\end{threeparttable}
\end{table}

\begin{table}[h]
	\centering
	\begin{threeparttable}
		\caption{Comparison of accuracies ($\%$) obtained by EFTS and random task combinations on \textit{mini}ImageNet, where the batchsize is 128. In addition, NUA = 50 and M = 4.}
		\begin{tabular}{cccc}
			\toprule[1pt]
			\textbf{}
			\bf{interval}        &{\bf{method}}&\bf{1 shot}&\bf{5 shot}\\
			\hline
			\multirow{2}{*}{every 5 epochs}   &Random	 &$57.00_{\pm 0.90}$ &$74.70_{\pm 0.69}$\\
			&EFTS	     &$57.87_{\pm 0.90}$ &$76.30_{\pm 0.28}$\\
			\hline
			\multirow{2}{*}{every 20 epochs}  &Random	 &$56.44_{\pm 0.88}$ &$75.34_{\pm 0.68}$\\
			&EFTS	     &$58.26_{\pm 0.87}$ &$76.92_{\pm 0.67}$\\
			\hline
			\multirow{2}{*}{once only}  &Random	&$57.66_{\pm 0.87}$ &$74.91_{\pm 0.69}$ \\
			&EFTS	     &$58.91_{\pm 0.88}$ &$76.90_{\pm 0.65}$\\
			\bottomrule[1pt]
		\end{tabular}\label{random}
	\end{threeparttable}
\end{table}

\begin{table*}[!htbp]
	\centering
	\setlength\tabcolsep{2pt}
	\begin{threeparttable}
		\caption{Accuracy ($\%$) comparison with the state-of-the-art for 5-way tasks on $\textit{mini}$ImageNet with 95$\%$ confidence intervals. $\ddag$ denotes that validation set as well as training set is used for training. For EFTS, NUA = 50 and M = 4.}
		\begin{tabular}{cccccccc}
			\toprule[1pt]
			
			\textbf{} 
			\multirow{2}{*}{\bf{Model}} & \multirow{2}{*}{\bf{Backbone}} &\multicolumn{2}{c}{\bf{\textit{mini}ImageNet}}
			&\multicolumn{2}{c}{\bf{\textit{tiered}-ImageNet}}
			&\multicolumn{2}{c}{\bf{CIFAR-FS}}\\
			\cmidrule(r){3-4}\cmidrule(r){5-6}\cmidrule(r){7-8}
			& &\bf{1-shot} &\bf{5-shot} 
			&\bf{1-shot} &\bf{5-shot}
			&\bf{1-shot} &\bf{5-shot}\\
			\hline
			Episodic methods\\
			\hline
			ProtoNet\cite{snell2017prototypical}           &ResNet-12         &$59.25_{\pm 0.64}$&$75.60_{\pm 0.48}$&$61.74_{\pm 0.77 }$   &$80.00_{\pm 0.55}$  &$72.2_{\pm 0.7 }$      &$83.5_{\pm 0.5 }$   \\ 
			TADAM\cite{oreshkin2018tadam}                &ResNet-12         &$58.50_{\pm 0.30}$   &$76.70_{\pm 0.30}$   &-&-&-&-\\
			AdaResNet\cite{munkhdalai2018rapid}           &ResNet-12         &$56.88_{\pm 0.62}$   &$71.94_{\pm 0.57 }$  &-&-&-&-\\
			LwoF\cite{gidaris2018dynamic}                 &WRN-28-10         &$60.06_{\pm 0.14 }$   &$76.39_{\pm 0.11 }$ &-&-&-&-  \\
			DSN\cite{simon2020adaptive}                   &ResNet-12         &$62.64_{\pm 0.66}$   &$78.83_{\pm 0.45}$  &$66.22_{\pm 0.75 }$   &$82.79_{\pm 0.48}$   &$72.3_{\pm 0.7 }$      &$85.1_{\pm 0.5 }$ \\  
			CTM\cite{li2019finding}                       &ResNet-18         &$62.05_{\pm 0.55}$   &$78.63_{\pm 0.06}$  &$64.78_{\pm 0.11}$   &$81.05_{\pm 0.52}$ &-&-\\
			Hyper ProtoNet\cite{Khrulkov2020}   &ResNet-18         &$59.47_{\pm 0.20}$   &$76.84_{\pm 0.14}$  &-&-&$64.02_{\pm 0.24}$&$82.53_{\pm 0.14}$\\	
			MetaOptNet-RR\cite{lee2019meta}      &ResNet-12         &$61.41_{\pm 0.61} $&$77.88_{\pm 0.46}$&$65.36_{\pm 0.71}$   &$81.34_{\pm 0.52}$ &$72.6_{\pm 0.7}$      &$84.3_{\pm 0.5}$ \\
			MetaOptNet-SVM\cite{lee2019meta}      &ResNet-12      &$62.64_{\pm 0.61}$   &$78.63_{\pm 0.46}$&$65.99_{\pm 0.72}$   &$81.56_{\pm 0.53}$  &$72.0_{\pm 0.7}$      &$84.2_{\pm 0.5}$ \\
			
			Ravichandran et al.\cite{ravichandran2019fewshot} &ResNet-12      &$59.04_{\pm - }$   &$77.64_{\pm -}$&$66.87_{\pm -}$   &$82.64_{\pm -}$  &$69.15_{\pm -}$ &$84.7_{\pm -}$\\
			CAN\cite{hou2019cross} &ResNet-12      &$\bm{63.85_{\pm 0.48 }}$   &$79.44_{\pm 0.34}$&$69.89_{\pm 0.51}$   &$84.23_{\pm 0.37}$  &-  &-\\
				\hline
				
			Non-episodic methods\\
			\hline
						Baseline\cite{chen2019closer}                 &ResNet-18         &$51.75_{\pm 0.80}$   &$74.27_{\pm 0.63}$  &-&-&$65.51_{\pm 0.87}$&$82.85_{\pm 0.55}$\\
			Baseline++\cite{chen2019closer}               &ResNet-18         &$51.87_{\pm 0.77}$   &$75.68_{\pm 0.63}$  &-&-&$67.02_{\pm 0.90}$&$83.58_{\pm 0.54}$\\
			NCA nearest centroid\cite{laenen2021episodes}  &ResNet-12 &$62.55_{\pm 0.12}$   &$78.27_{\pm 0.09}$ &$68.35_{\pm 0.13}$      &$83.20_{\pm 0.10}$ &$72.49_{\pm 0.12}$   &$85.15_{\pm 0.09}$   \\
            RFIC-simple \cite{tian2020rethinking} &ResNet-12 &$62.02_{\pm 0.63}$   &$79.64_{\pm 0.44}$ &$69.74_{\pm 0.72}$ &$84.41_{\pm 0.55}$ &$71.5_{\pm 0.8}$   &$86.0_{\pm 0.5}$   \\ 
            						\hline 
Our episodic-free methods\\
\hline
			EFTS(ProtoNet for target tasks)                      &ResNet-12         &$61.77_{\pm 0.89}$   &$79.27_{\pm 0.62}$ &$69.46_{\pm 0.96}$&$84.30_{\pm 0.66}$&$72.92_{\pm 0.94}$&$85.74_{\pm 0.64}$  \\
			EFTS(LR for target tasks)                      &ResNet-12         &$63.77_{\pm 0.85}$   &$\bm{79.82_{\pm 0.55}}$ &$\bm{70.11_{\pm 0.89}}$&$\bm{84.89_{\pm 0.64}}$&$\bm{74.85_{\pm 0.84}}$&$\bm{87.41_{\pm 0.59}}$  \\
			\bottomrule[1pt]
			
		\end{tabular}\label{SOTA}
	\end{threeparttable}
\end{table*}

As well as the strategy of all task selection, random task selection is also simple and would be effective. The random task selection can be repeated during the training process. Table \ref{random} shows the performance comparison of EFTS and random task selection with different interval of task selection on \textit{mini}ImageNet . The results in Table \ref{random} shows that Whether the task is selected once or multiple times, selecting tasks using the EFTS criterion is better than randomly selecting tasks for 1 shot and 5 shot tasks. Although in the worst-case scenario, the accuracy obtained by random task selection is only $2\%$ lower than that of EFTS, its performance depends on the task types in the task set. That is to say, when more tasks in the task set are unhelpful, the performance of random task selection may be worse.

\subsection{Comparison with State-of-the-art}

In the section, we report our results based on ResNet-12 and compare them with two kinds of recent representative methods that includes episodic training methods and non-episodic training methods on three commonly used datasets, as shown Table \ref{SOTA}. Here we adopt ProtoNet head and LR head for comparison. For protoNet head, we employed the strategy of sharing inputs, while for LR head, we utilized a strategy of not sharing inputs. In comparison of \cite{tian2020rethinking}, we did not compare against results after knowledge distillation because it is an embeddable method that can be applied to most methods including ours. In addition, for ProtoNet, the way is 16, and batchsize is 256; for LR head, the batchsize is 64. For all the heads, we applied data augmentation on the inputs for task1-task5. In Table \ref{SOTA}, we can see that EFTS with LR head outperforms ProtoNet head and other methods, which is benefited from task4 (classification) that is proposed in \cite{tian2020rethinking}. Nevertheless, these results of the proposed EFTS with ProtoNet head is comparable to \cite{laenen2021episodes} that applied a greater batchsize. Overall, these results show that task selection strategies are competitive to the conventional fixed task strategies.

It should be noted that our goal is not to improve the state-of-the-art performance, but rather to demonstrate the practicality of EFTS in the context of related methods. In light of this aim, and to render our approach more comprehensible, we choice the often used methods rather than those with superior performance but not yet representative as the candidates in the task set. This choice may limit the performance of EFTS, and thus we did not employ more methods for comparison. Nonetheless, a further improved performance can be achieved by adding some stunning approaches (such as \cite{Yang2022few1, kang2021relational, zhang2020deepemd, dhillon2020baseline}) in the task set .

%% file: 6_discussion.tex
\section{Is validation data selection critical?}

Intuitively, different tasks may be selected when using cross-domain data for validation. Moreover, if the distribution of samples used for validation is different from that of test samples, the selection of tasks deviates from the target and would have a negative impact on the model's performance in the target task. Here we will discuss whether the selection of validation sets is crucial. In Table \ref{Crossdomain}, we compare the performance respectively using  CIFAR-FS and \textit{mini}ImageNet for evaluation while using CIFAR-FS for training. Counterintuitively, we see that the performance of the model is not sensitive to the selection of validation samples, regardless of whether multiple task selections were made. Similar results are hold in \cite{fifty2021efficiently}, which shows that the inter-task affinity scores computed on the validation set is highly similar to those computed on the training set. Nevertheless, we still need to emphasize that these results does not deny the importance of validating data selection in all the cases.

\begin{table}[h]
	\centering
	\begin{threeparttable}
		\caption{Train-data of CIFAR-FS for training and different data sources including train-data of CIFAR-FS, evaluation-data of CIFAR-FS and \textit{mini}ImageNet for evaluating, where Q = 5 and batchsize is 128. In addition, NUA = 50 and M = 4.}
		\begin{tabular}{cccc}
			\toprule[1pt]
			\textbf{}
			\bf{interval} &{\bf{evaluation source}}&\bf{1 shot}&\bf{5 shot}\\
			\hline
			\multirow{2}{*}{once only} &\textit{mini}ImageNet          &$67.30_{\pm 0.98}$ &$82.18_{\pm 0.71}$\\
			&CIFAR-FS	                      &$68.18_{\pm 0.95}$ &$82.87_{\pm 0.72}$\\
			\hline
			every 5 &\textit{mini}ImageNet	  &$68.01_{\pm 0.93}$ &$82.88_{\pm 0.71}$\\
			epochs &CIFAR-FS	                      &$67.56_{\pm 0.97}$ &$82.60_{\pm 0.70}$\\
			\bottomrule[1pt]
		\end{tabular}\label{Crossdomain}
	\end{threeparttable}
\end{table}

%% file: 7_conclusion.tex
\section{Conclusion and Future Work}

In this work, we propose a framework called EFTS for few-shot learning. EFTS supports effective episodic-free tasks for training meta-learners, which thus expands the episodic training paradigm typically used for few-shot learning. In EFTS, the episodic-free tasks are selected as the training tasks from a task set, according to the their benefits on a series of episodic tasks. We analyzed the impact of its hyper-parameters and demonstrated the effectiveness of EFTS through ablation experiments and comparison experiments with state-of-the-art methods.

There are two directions worth studying in future work. Firstly, the evaluation for task selection can be developed to make more accurate predictions of long-term impacts of a task subset on the performance of meta-learners. Secondly, our work makes it possible to more effectively select multiple tasks for few-shot learning. The task set including homogeneous and heterogeneous tasks used for task selection can be improved in future work through task engineering.

%% file: main.bbl
\begin{thebibliography}{47}
\providecommand{\natexlab}[1]{#1}
\providecommand{\url}[1]{\texttt{#1}}
\expandafter\ifx\csname urlstyle\endcsname\relax
  \providecommand{\doi}[1]{doi: #1}\else
  \providecommand{\doi}{doi: \begingroup \urlstyle{rm}\Url}\fi

\bibitem[Agrawal et~al.(2021)Agrawal, Abutarboush, Ganesh, and
  Mohamed]{agrawal2021metaheuristic}
Prachi Agrawal, Hattan~F Abutarboush, Talari Ganesh, and Ali~Wagdy Mohamed.
\newblock Metaheuristic algorithms on feature selection: A survey of one decade
  of research (2009-2019).
\newblock \emph{Ieee Access}, 9:\penalty0 26766--26791, 2021.

\bibitem[Allen et~al.(2019)Allen, Shelhamer, Shin, and
  Tenenbaum]{allen2019infinite}
Kelsey Allen, Evan Shelhamer, Hanul Shin, and Joshua Tenenbaum.
\newblock Infinite mixture prototypes for few-shot learning.
\newblock In \emph{International conference on machine learning}, pages
  232--241. PMLR, 2019.

\bibitem[Altae-Tran et~al.(2017)Altae-Tran, Ramsundar, Pappu, and
  Pande]{altae2017low}
Han Altae-Tran, Bharath Ramsundar, Aneesh~S Pappu, and Vijay Pande.
\newblock Low data drug discovery with one-shot learning.
\newblock \emph{ACS central science}, 3\penalty0 (4):\penalty0 283--293, 2017.

\bibitem[Awadallah et~al.(2020)Awadallah, Al-Betar, Hammouri, and
  Alomari]{awadallah2020binary}
Mohammed~A Awadallah, Mohammed~Azmi Al-Betar, Abdelaziz~I Hammouri, and
  Osama~Ahmad Alomari.
\newblock Binary jaya algorithm with adaptive mutation for feature selection.
\newblock \emph{Arabian Journal for Science and Engineering}, 45\penalty0
  (12):\penalty0 10875--10890, 2020.

\bibitem[Bertinetto et~al.(2018)Bertinetto, Henriques, Torr, and
  Vedaldi]{bertinetto2018meta}
Luca Bertinetto, Joao~F Henriques, Philip~HS Torr, and Andrea Vedaldi.
\newblock Meta-learning with differentiable closed-form solvers.
\newblock \emph{arXiv preprint arXiv:1805.08136}, 2018.

\bibitem[Bouniot et~al.(2022)Bouniot, Redko, Audigier, Loesch, and
  Habrard]{bouniot2022improving}
Quentin Bouniot, Ievgen Redko, Romaric Audigier, Ang{\'e}lique Loesch, and
  Amaury Habrard.
\newblock Improving few-shot learning through multi-task representation
  learning theory.
\newblock In \emph{European Conference on Computer Vision}, pages 435--452.
  Springer, 2022.

\bibitem[Chen et~al.(2020)Chen, Wu, Li, Li, Zhan, and Chung]{chen2020closer}
Jiaxin Chen, Xiao-Ming Wu, Yanke Li, Qimai Li, Li-Ming Zhan, and Fu-lai Chung.
\newblock A closer look at the training strategy for modern meta-learning.
\newblock \emph{Advances in Neural Information Processing Systems},
  33:\penalty0 396--406, 2020.

\bibitem[Chen et~al.(2019)Chen, Liu, Kira, Wang, and Huang]{chen2019closer}
Wei-Yu Chen, Yen-Cheng Liu, Zsolt Kira, Yu-Chiang~Frank Wang, and Jia-Bin
  Huang.
\newblock A closer look at few-shot classification.
\newblock \emph{arXiv preprint arXiv:1904.04232}, 2019.

\bibitem[Chen et~al.(2021)Chen, Liu, Xu, Darrell, and Wang]{chen2021meta}
Yinbo Chen, Zhuang Liu, Huijuan Xu, Trevor Darrell, and Xiaolong Wang.
\newblock Meta-baseline: Exploring simple meta-learning for few-shot learning.
\newblock In \emph{Proceedings of the IEEE/CVF international conference on
  computer vision}, pages 9062--9071, 2021.

\bibitem[Dhillon et~al.(2020)Dhillon, Chaudhari, Ravichandran, and
  Soatto]{dhillon2020baseline}
Guneet~S. Dhillon, Pratik Chaudhari, Avinash Ravichandran, and Stefano Soatto.
\newblock A baseline for few-shot image classification.
\newblock In \emph{International Conference on Learning Representations}, 2020.

\bibitem[Du et~al.(2020)Du, Hu, Kakade, Lee, and Lei]{du2020few}
Simon~S Du, Wei Hu, Sham~M Kakade, Jason~D Lee, and Qi Lei.
\newblock Few-shot learning via learning the representation, provably.
\newblock \emph{arXiv preprint arXiv:2002.09434}, 2020.

\bibitem[Fei-Fei et~al.(2006)Fei-Fei, Fergus, and Perona]{fei2006one}
Li Fei-Fei, Robert Fergus, and Pietro Perona.
\newblock One-shot learning of object categories.
\newblock \emph{IEEE transactions on pattern analysis and machine
  intelligence}, 28\penalty0 (4):\penalty0 594--611, 2006.

\bibitem[Fifty et~al.(2021)Fifty, Amid, Zhao, Yu, Anil, and
  Finn]{fifty2021efficiently}
Chris Fifty, Ehsan Amid, Zhe Zhao, Tianhe Yu, Rohan Anil, and Chelsea Finn.
\newblock Efficiently identifying task groupings for multi-task learning.
\newblock \emph{Advances in Neural Information Processing Systems},
  34:\penalty0 27503--27516, 2021.

\bibitem[Finn et~al.(2017)Finn, Abbeel, and Levine]{finn2017model}
Chelsea Finn, Pieter Abbeel, and Sergey Levine.
\newblock Model-agnostic meta-learning for fast adaptation of deep networks.
\newblock In \emph{International conference on machine learning}, pages
  1126--1135. PMLR, 2017.

\bibitem[Fort(2017)]{fort2017gaussian}
Stanislav Fort.
\newblock Gaussian prototypical networks for few-shot learning on omniglot.
\newblock \emph{arXiv preprint arXiv:1708.02735}, 2017.

\bibitem[Gidaris and Komodakis(2018)]{gidaris2018dynamic}
Spyros Gidaris and Nikos Komodakis.
\newblock Dynamic few-shot visual learning without forgetting.
\newblock In \emph{Proceedings of the IEEE Conference on Computer Vision and
  Pattern Recognition}, pages 4367--4375, 2018.

\bibitem[Hospedales et~al.(2021)Hospedales, Antoniou, Micaelli, and
  Storkey]{hospedales2021meta}
Timothy Hospedales, Antreas Antoniou, Paul Micaelli, and Amos Storkey.
\newblock Meta-learning in neural networks: A survey.
\newblock \emph{IEEE transactions on pattern analysis and machine
  intelligence}, 44\penalty0 (9):\penalty0 5149--5169, 2021.

\bibitem[Hou et~al.(2019)Hou, Chang, Ma, Shan, and Chen]{hou2019cross}
R. Hou, H. Chang, B. Ma, S. Shan, and X. Chen.
\newblock Cross attention network for few-shot classification.
\newblock In \emph{NIPS}, pages 4005--4016, 2019.

\bibitem[Kang et~al.(2021)Kang, Kwon, Min, and Cho]{kang2021relational}
D. Kang, H. Kwon, J. Min, and M. Cho.
\newblock Relational embedding for few-shot classification.
\newblock In \emph{ICCV}, pages 8822--8833, 2021.

\bibitem[Kendall et~al.(2018)Kendall, Gal, and Cipolla]{kendall2018multi}
Alex Kendall, Yarin Gal, and Roberto Cipolla.
\newblock Multi-task learning using uncertainty to weigh losses for scene
  geometry and semantics.
\newblock In \emph{Proceedings of the IEEE conference on computer vision and
  pattern recognition}, pages 7482--7491, 2018.

\bibitem[Khosla et~al.(2020)Khosla, Teterwak, Wang, Sarna, Tian, Isola,
  Maschinot, Liu, and Krishnan]{khosla2020supervised}
Prannay Khosla, Piotr Teterwak, Chen Wang, Aaron Sarna, Yonglong Tian, Phillip
  Isola, Aaron Maschinot, Ce Liu, and Dilip Krishnan.
\newblock Supervised contrastive learning.
\newblock \emph{Advances in neural information processing systems},
  33:\penalty0 18661--18673, 2020.

\bibitem[Khrulkov et~al.(2020)Khrulkov, Mirvakhabova, Ustinova, and
  Oseledets]{Khrulkov2020}
V. Khrulkov, L. Mirvakhabova, E. Ustinova, and I. Oseledets.
\newblock Hyperbolic image embeddings.
\newblock In \emph{Proceedings of the IEEE/CVF Conference on Computer Vision
  and Pattern Recognition}, pages 6418–--6428, 2020.

\bibitem[Krizhevsky et~al.(2009)Krizhevsky, Hinton,
  et~al.]{krizhevsky2009learning}
Alex Krizhevsky, Geoffrey Hinton, et~al.
\newblock Learning multiple layers of features from tiny images.
\newblock 2009.

\bibitem[Laenen and Bertinetto(2021)]{laenen2021episodes}
Steinar Laenen and Luca Bertinetto.
\newblock On episodes, prototypical networks, and few-shot learning.
\newblock \emph{Advances in Neural Information Processing Systems},
  34:\penalty0 24581--24592, 2021.

\bibitem[Lee et~al.(2019)Lee, Maji, Ravichandran, and Soatto]{lee2019meta}
Kwonjoon Lee, Subhransu Maji, Avinash Ravichandran, and Stefano Soatto.
\newblock Meta-learning with differentiable convex optimization.
\newblock In \emph{Proceedings of the IEEE/CVF conference on computer vision
  and pattern recognition}, pages 10657--10665, 2019.

\bibitem[Leung et~al.(2021)Leung, Fung, and Hoi]{leung2021health}
Carson~K Leung, Daryl~LX Fung, and Calvin~SH Hoi.
\newblock Health analytics on covid-19 data with few-shot learning.
\newblock In \emph{Big Data Analytics and Knowledge Discovery: 23rd
  International Conference, DaWaK 2021, Virtual Event, September 27--30, 2021,
  Proceedings 23}, pages 67--80. Springer, 2021.

\bibitem[Li et~al.(2019)Li, Eigen, Dodge, Zeiler, and Wang]{li2019finding}
Hongyang Li, David Eigen, Samuel Dodge, Matthew Zeiler, and Xiaogang Wang.
\newblock Finding task-relevant features for few-shot learning by category
  traversal.
\newblock In \emph{Proceedings of the IEEE/CVF conference on computer vision
  and pattern recognition}, pages 1--10, 2019.

\bibitem[Liu et~al.(2020)Liu, Song, and Qin]{liu2020prototype}
Jinlu Liu, Liang Song, and Yongqiang Qin.
\newblock Prototype rectification for few-shot learning.
\newblock In \emph{Computer Vision--ECCV 2020: 16th European Conference,
  Glasgow, UK, August 23--28, 2020, Proceedings, Part I 16}, pages 741--756.
  Springer, 2020.

\bibitem[Munkhdalai et~al.(2018)Munkhdalai, Yuan, Mehri, and
  Trischler]{munkhdalai2018rapid}
Tsendsuren Munkhdalai, Xingdi Yuan, Soroush Mehri, and Adam Trischler.
\newblock Rapid adaptation with conditionally shifted neurons.
\newblock In \emph{International Conference on Machine Learning}, pages
  3664--3673, 2018.

\bibitem[Oreshkin et~al.(2018)Oreshkin, Rodr{\'\i}guez~L{\'o}pez, and
  Lacoste]{oreshkin2018tadam}
Boris Oreshkin, Pau Rodr{\'\i}guez~L{\'o}pez, and Alexandre Lacoste.
\newblock Tadam: Task dependent adaptive metric for improved few-shot learning.
\newblock \emph{Advances in neural information processing systems}, 31, 2018.

\bibitem[Pahde et~al.(2021)Pahde, Puscas, Klein, and Nabi]{pahde2021multimodal}
Frederik Pahde, Mihai Puscas, Tassilo Klein, and Moin Nabi.
\newblock Multimodal prototypical networks for few-shot learning.
\newblock In \emph{Proceedings of the IEEE/CVF Winter Conference on
  Applications of Computer Vision}, pages 2644--2653, 2021.

\bibitem[Paszke et~al.(2017)Paszke, Gross, Chintala, Chanan, Yang, DeVito, Lin,
  Desmaison, Antiga, and Lerer]{paszke2017automatic}
Adam Paszke, Sam Gross, Soumith Chintala, Gregory Chanan, Edward Yang, Zachary
  DeVito, Zeming Lin, Alban Desmaison, Luca Antiga, and Adam Lerer.
\newblock Automatic differentiation in pytorch.
\newblock 2017.

\bibitem[Raghu et~al.(2019)Raghu, Raghu, Bengio, and Vinyals]{raghu2019rapid}
Aniruddh Raghu, Maithra Raghu, Samy Bengio, and Oriol Vinyals.
\newblock Rapid learning or feature reuse? towards understanding the
  effectiveness of maml.
\newblock \emph{arXiv preprint arXiv:1909.09157}, 2019.

\bibitem[Ravichandran et~al.(2019)Ravichandran, Bhotika, and
  Soatto]{ravichandran2019fewshot}
A. Ravichandran, R. Bhotika, and S. Soatto.
\newblock Few-shot learning with embedded class models and shot-free meta
  training.
\newblock In \emph{ICCV}, pages 331--339, 2019.

\bibitem[Ren et~al.(2018)Ren, Triantafillou, Ravi, Snell, Swersky, Tenenbaum,
  Larochelle, and Zemel]{ren2018meta}
Mengye Ren, Eleni Triantafillou, Sachin Ravi, Jake Snell, Kevin Swersky,
  Joshua~B Tenenbaum, Hugo Larochelle, and Richard~S Zemel.
\newblock Meta-learning for semi-supervised few-shot classification.
\newblock \emph{arXiv preprint arXiv:1803.00676}, 2018.

\bibitem[Sener and Koltun(2018)]{sener2018multi}
Ozan Sener and Vladlen Koltun.
\newblock Multi-task learning as multi-objective optimization.
\newblock \emph{Advances in neural information processing systems}, 31, 2018.

\bibitem[Simon et~al.(2020)Simon, Koniusz, Nock, and
  Harandi]{simon2020adaptive}
Christian Simon, Piotr Koniusz, Richard Nock, and Mehrtash Harandi.
\newblock Adaptive subspaces for few-shot learning.
\newblock In \emph{Proceedings of the IEEE/CVF conference on computer vision
  and pattern recognition}, pages 4136--4145, 2020.

\bibitem[Snell et~al.(2017)Snell, Swersky, and Zemel]{snell2017prototypical}
Jake Snell, Kevin Swersky, and Richard Zemel.
\newblock Prototypical networks for few-shot learning.
\newblock \emph{Advances in neural information processing systems}, 30, 2017.

\bibitem[Sung et~al.(2018)Sung, Yang, Zhang, Xiang, Torr, and
  Hospedales]{sung2018learning}
Flood Sung, Yongxin Yang, Li Zhang, Tao Xiang, Philip~HS Torr, and Timothy~M
  Hospedales.
\newblock Learning to compare: Relation network for few-shot learning.
\newblock In \emph{Proceedings of the IEEE conference on computer vision and
  pattern recognition}, pages 1199--1208, 2018.

\bibitem[Tian et~al.(2020)Tian, Wang, Krishnan, Tenenbaum, and
  Isola]{tian2020rethinking}
Yonglong Tian, Yue Wang, Dilip Krishnan, Joshua~B Tenenbaum, and Phillip Isola.
\newblock Rethinking few-shot image classification: a good embedding is all you
  need?
\newblock In \emph{Computer Vision--ECCV 2020: 16th European Conference,
  Glasgow, UK, August 23--28, 2020, Proceedings, Part XIV 16}, pages 266--282.
  Springer, 2020.

\bibitem[Upadhyay et~al.(2023)Upadhyay, Chhipa, Phlypo, Saini, and
  Liwicki]{upadhyay2023multi}
Richa Upadhyay, Prakash~Chandra Chhipa, Ronald Phlypo, Rajkumar Saini, and
  Marcus Liwicki.
\newblock Multi-task meta learning: learn how to adapt to unseen tasks.
\newblock In \emph{2023 International Joint Conference on Neural Networks
  (IJCNN)}, pages 1--10. IEEE, 2023.

\bibitem[Vinyals et~al.(2016)Vinyals, Blundell, Lillicrap, Kavukcuoglu, and
  Wierstra]{vinyals2016matching}
Oriol Vinyals, Charles Blundell, Timothy Lillicrap, Koray Kavukcuoglu, and Daan
  Wierstra.
\newblock Matching networks for one shot learning.
\newblock In \emph{Advances in neural information processing systems}, pages
  3637--3645, 2016.

\bibitem[Wang et~al.(2020)Wang, Yao, Kwok, and Ni]{wang2020generalizing}
Yaqing Wang, Quanming Yao, James~T Kwok, and Lionel~M Ni.
\newblock Generalizing from a few examples: A survey on few-shot learning.
\newblock \emph{ACM computing surveys (csur)}, 53\penalty0 (3):\penalty0 1--34,
  2020.

\bibitem[Yang et~al.(2022{\natexlab{a}})Yang, Guo, Li, Marinello, Ercisli, and
  Zhang]{yang2022survey}
Jiachen Yang, Xiaolan Guo, Yang Li, Francesco Marinello, Sezai Ercisli, and
  Zhuo Zhang.
\newblock A survey of few-shot learning in smart agriculture: developments,
  applications, and challenges.
\newblock \emph{Plant Methods}, 18\penalty0 (1):\penalty0 1--12,
  2022{\natexlab{a}}.

\bibitem[Yang et~al.(2022{\natexlab{b}})Yang, Wang, and Zhu]{yang2022few}
Zhanyuan Yang, Jinghua Wang, and Yingying Zhu.
\newblock Few-shot classification with contrastive learning.
\newblock In \emph{European Conference on Computer Vision}, pages 293--309.
  Springer, 2022{\natexlab{b}}.

\bibitem[Zhang et~al.(2020)Zhang, Cai, Lin, and Shen]{zhang2020deepemd}
C. Zhang, Y. Cai, G. Lin, and C. Shen.
\newblock Deepemd: Few-shot image classification with differentiable earth
  mover's distance and structured classifiers.
\newblock In \emph{CVPR}, pages 12200--12210, 2020.

\bibitem[Zhanyuan et~al.(2022)Zhanyuan, Jinghua, and Yingying]{Yang2022few1}
Yang Zhanyuan, Wang Jinghua, and Zhu Yingying.
\newblock Few-shot classification with contrastive learning.
\newblock In \emph{ECCV(2022)}, pages 1--14, 2022.

\end{thebibliography}
